\documentclass[conference]{IEEEtran}
\IEEEoverridecommandlockouts

\usepackage{cite}

\usepackage[absolute,overlay]{textpos}
\usepackage{fancyhdr}
\usepackage{lipsum} 
\usepackage{amsmath,amssymb,amsfonts}
\usepackage{algorithmic}
\usepackage{graphicx}
\usepackage{textcomp}
\usepackage{xcolor}
\def\BibTeX{{\rm B\kern-.05em{\sc i\kern-.025em b}\kern-.08em
    T\kern-.1667em\lower.7ex\hbox{E}\kern-.125emX}}

\begin{document}

\title{Tweedie Regression for Video Recommendation System}

\author{\IEEEauthorblockN{1\textsuperscript{st} Yan Zheng}
\IEEEauthorblockA{\textit{Tubi}\\
Beijing, China \\
yanzheng@tubi.tv}
\and
\IEEEauthorblockN{2\textsuperscript{nd} Qiang Chen}
\IEEEauthorblockA{\textit{Tubi}\\
Sunnyvale, United States \\
qiang@tubi.tv}
\and
\IEEEauthorblockN{3\textsuperscript{rd} Chenglei Niu}
\IEEEauthorblockA{\textit{Tubi}\\
San Francisco, United States \\
chenglei@tubi.tv}
}

\maketitle

\begin{textblock*}{\textwidth}(0cm,26cm) 
  \centering
  \footnotesize
  \copyright~2025 IEEE. Personal use of this material is permitted. Permission from IEEE must be obtained for all other uses, in any current or future media, including reprinting/republishing this material for advertising or promotional purposes, creating new collective works, for resale or redistribution to servers or lists, or reuse of any copyrighted component of this work in other works.
\end{textblock*}

\pagestyle{fancy}
\fancyhf{} 
\fancyfoot[C]{\footnotesize \copyright~2025 IEEE. Personal use of this material is permitted. 
Permission from IEEE must be obtained for all other uses, in any current or future media, 
including reprinting/republishing this material for advertising or promotional purposes, 
creating new collective works, for resale or redistribution to servers or lists, 
or reuse of any copyrighted component of this work in other works.}

\setlength{\footskip}{20pt}  

\begin{abstract}

Modern recommendation systems aim to increase click-through rates (CTR) for better user experience, through commonly treating ranking as a classification task focused on predicting CTR. However, there's a gap between this method and the actual objectives of businesses across different sectors. In video recommendation services, the objective of video on demand (VOD) extends beyond merely encouraging clicks, but also guiding users to discover their true interests, leading to increased watch time. And longer users' watch time will leads to more revenue through increased chances of presenting online display advertisements. This research addresses the issue by redefining the problem from classification to regression, with a focus on maximizing revenue through user viewing time. Due to the lack of positive labels on recommendation, the study introduces Tweedie Loss Function, which is better suited in this scenario than the traditional mean square error loss. The paper also provides insights on how Tweedie process capture users' diverse interests. Our offline simulation and online A/B test revealed that we can substantially enhance our core business objectives: user engagement in terms of viewing time and, consequently, revenue. Additionally, we provide a theoretical comparison between the Tweedie Loss and the commonly employed viewing time weighted Logloss, highlighting why Tweedie Regression stands out as an efficient solution. We further outline a framework for designing a loss function that focuses on a singular objective.

\end{abstract}


\section{Introduction}

Common recommendation system generally consists of two key elements: the recaller and the ranker. The recaller selects a shortlist of potential candidates from a vast pool. The ranker then is responsible for ranking items within this set before presenting to audiences. In the video-on-demand(VOD) industry, effective recommendation systems are critical for improving user experience by helping viewers find preferred content or explore new interests\cite{YouTubeVideoRec, DNYoutubeRec}. Unlike some scenarios like e-commerce, where the focus is on optimizing CTR, VOD aims at maximizing overall viewing time because it increases the chance to show more online advertisements which leads to greater revenue.
Given the design of recommendation systems, it is the ranker's task to prioritize content that is likely to hold users' attention for longer periods. We believe the ranker model focused on optimizing total viewing time is an industrial challenge. We propose a solution that ranks content based on predicted viewing times by a model built on Tweedie regression.

\begin{figure}[h]
	\label{tweedie_dist}
	\centering
	\includegraphics[width=0.5\textwidth]{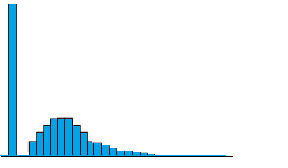}
	\small
	\caption{Typical Tweedie Distribution pattern with p=1.5, this is the histogram of x where x follows Tweedie Distribution}
\end{figure}

The Tweedie distribution is a type of compound Poisson-gamma distribution that has positive mass at zero, but is otherwise continuous. It also has a peak at some points above zero. Figure \ref{tweedie_dist} shows what the distribution of Tweedie looks like. It is a subfamily of reproductive exponential dispersion models (EDMs) with a unique mean-variance relationship. The hyper-parameter called power parameter, $p$, links the mean and variance following the relationship $Var(x)=\phi \mu^p$, where $\mu$ and $\phi$ are mean \& dispersion parameter respectively. Different $p$ specifies which the sub-class of distributions within EDMs family.

\begin{figure}[h]
	\label{normalized_vt}
	\centering
	\includegraphics[width=0.4\textwidth]{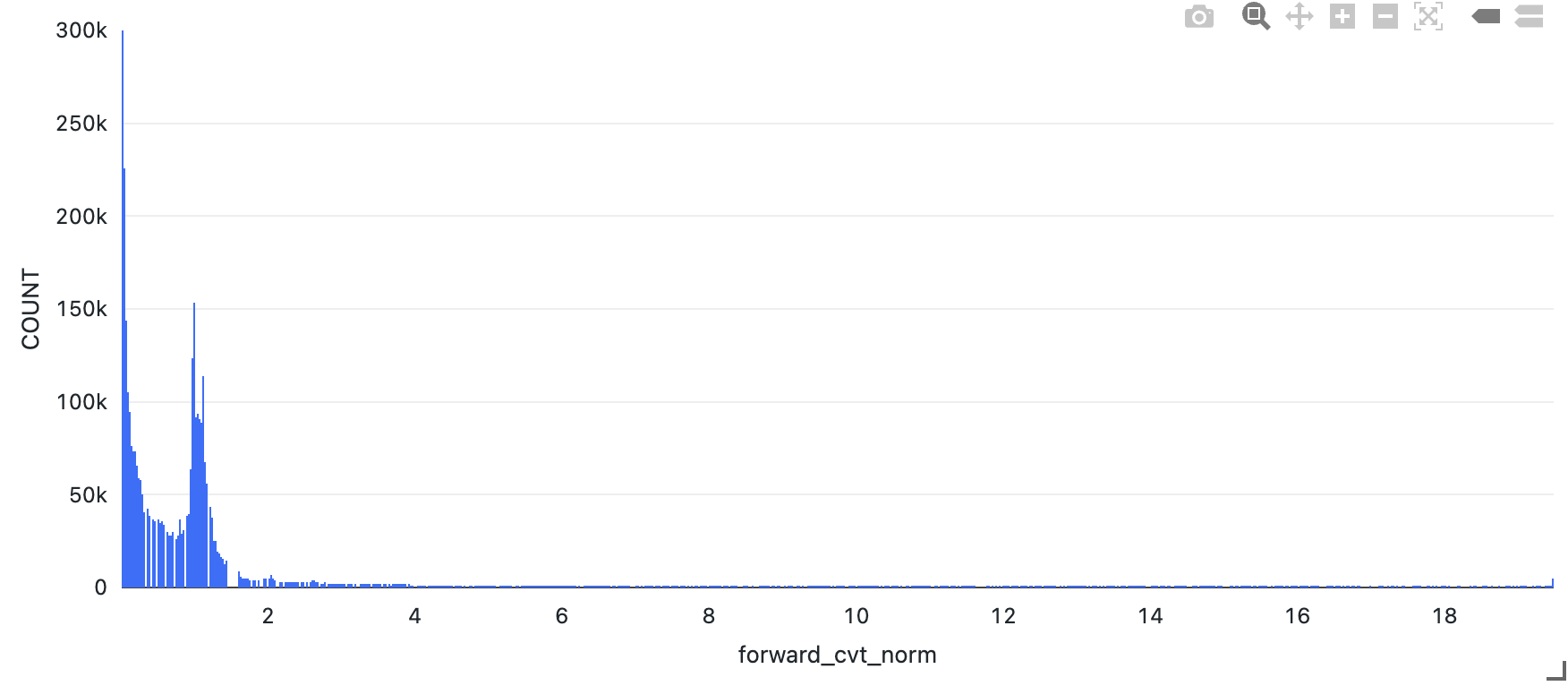}
	\small
	\caption{Real-world user-item pairs' watch time on Tubi over a given period, this is also the histogram of watch times}
\end{figure}

Comparatively, figure \ref{normalized_vt} presents the normalized real-world watch time distribution for each content across all presentations. It is evident that the watch time distribution is not Gaussian but closely resembles a Tweedie distribution, peak at zero and smooth beyond zero. Meanwhile, extensive real-world analyses of users' watch time distributions, such as \cite{NoisedWatchTime}, suggest that the Tweedie distribution best aligns with real-world patterns.

More specifically, Tweedie distribution is generated by the so called Tweedie process, which is an aggregation of several sub-processes. A random variable which follows Tweedie process can be characterized as

\begin{equation}
	X = 
	\begin{cases}
		0 & M=0 \\
		\sum_{i=1}^{M} C_i & M>0 \\
	\end{cases}  
\end{equation} 
where $M$ follows Possion and $C_i$ follows Gamma process with the same mean \& dispersion parameters, and $M$ is independent of $C_i$. Put differently, a Poisson process initially dictates the number of events, $M$, that occur, followed by a Gamma process for each event that defines its magnitude or intensity, $C_i$. The final outcome is the aggregation of $C_i$ of all the events.\cite{enwiki:1220054721} 

The probability function of Tweedie distribution can be written as
\begin{equation}
	f(x|\mu, \phi, p) = \alpha(x, \phi, p) \cdot \exp \{ \frac{1}{\phi} x \cdot \frac{\mu^{1-p}}{1-p} - \frac{\mu^{2-p}}{2-p} ) \} 
\end{equation}
where $\alpha(x, \phi, p)$ is a normalizing constant that ensures this is a valid probability density function. For this type of exponential dispersion model family, the normalized factor can be solved by the integration equation, $\int f(x|\mu, \phi, p) = 1$. 

In ordinary linear regression, it is assumed that the residuals of the target variables exhibit a Gaussian distribution, and hence minimizing the mean square error loss is equivalent to maximizing the likelihood function. Similarly, if the variability of the target variable follows the Tweedie distribution, the corresponding loss function, which can be derived from maximizing likelihood, is:

\begin{equation} \label{eq:2}
	- \sum_{i} x_i \cdot \frac{\widetilde{x_i}^{1-p}}{1-p} + \frac{\widetilde{x_i}^{2-p}}{2-p}
\end{equation}
where $\widetilde{x_i}$ is the prediction and $x_i$ is the real value.

Our key contributions include a theoretical analysis of Tweedie loss and log-loss within Hilbert space and the development of a framework for combining different loss functions. Additionally, we present detailed results from offline simulations and online experiments to validate the effectiveness of our proposed method.

The paper is constructed into several parts: Section II reviews past research on Tweedie model within insurance society, loss functions based on learning-to-rank framework, as well as multi-interests in recommendation systems. Section III provides comparison analyses between Tweedie regression and the conventional Logloss method. It also includes a strategy for achieving a more dedicated objective by weighted summation of several different objectives. Section IV presents its performance in offline simulation and online A/B tests which shows that Tweedie regression is more suitable for optimizing viewing time and revenue. The final section summarizes our works and suggests directions for future research. 

\section{Related Works}

\subsection{Tweedie in insurance}

The Tweedie regression model has been extensively used in the insurance industry for claim modeling. When underwriting an insurance policy, companies are tasked on predicting the expected costs that may arise during policy term, and use that estimate to efficient pricing. A noteworthy characteristic of these real world costs is that the majority of policyholders may not file any claims at all, resulting in a distribution of claim amounts that is predominantly concentrated at zero and otherwise continuous. This asymmetric distributional pattern bears a resemblance to that of the Tweedie distribution\cite{TweedieInsurance}. Actuarial sciences continue doing research in this area for enhancement of insurance pricing models. For example, Denuit\cite{DENUIT2021485} have explored the incorporation of autocalibration techniques aimed at correcting biases in this form of generalized linear model. Donatien\cite{Hainaut26112022} shows that boosting can be conducted directly on the response under Tweedie loss function, which attracts attentions among actuaries. And in terms of Tweedie model enhancement, Qian etl\cite{Qian02042016}. analyzed the elastic net on Tweedie's compound poisson model. Meanwhile, Double Poisson-Tweedie regression model\cite{PetterleBonatKokonendjiSeganfredoMoraesdaSilva+2019} has been proposed in medical research also for studying the HIV infections. 

\subsection{Learning to rank}

The design of loss functions for ranking problems remains a hot research topic within the machine learning community. Historically, ranking problem have been treated as a type of classification problem, typically applying Logloss for point-wise rankers or utilizing the comparison of positive and negative pairs for pair-wise rankers\cite{Burges2005}. Subsequently, list-wise loss functions have been introduced aiming at optimizing the entire list of items to achieve an ideal ordering, a concept pioneered by Hang Li\cite{ListwiseRank}, and still remains an active area of research to this day\cite{8356920, 9388879}. A later research.\cite{burges2010from} within the learning-to-rank domain has emerged at incorporating elements from information retrieval performance metrics like NDCG into more sophisticated loss functions.

However, these efforts primarily target the optimization of CTR. Until more recently, Google's introduced \cite{DNYoutubeRec} a modern recommendation system utilizing deep learning that aimed at optimizing user viewing time. By applying video watch time as weights of samples during the training phase, they argue that the learned model could approximately maximizing viewing time.

More recently, there are also numerous studies\cite{Wang2020CapturingAD, NoisedWatchTime, dwellTimeForPersonalization, TreebasedProgressiveReg} that concentrate on watch time prediction for short-video recommendations due to the scarcity of reliable positive labels in business like Tiktok or Kuaishou. However, these studies have not explicitly explored the context of long-form video recommendations, such as movies or series. We've also observed that certain techniques within short-video recommendation domain are not transferable to long video VOD business through our internal experiment system.

\subsection{Multi-Interests}

Concurrently, a substantial amount of research has been conducted on multi-interest or multi-user recommendation systems, which provides the foundation for our approach of the alternative regression—Tweedie regression. Previous work, such as Murphy's study\cite{subspace_clustering} on multi-user identification, has found a range of applications in real-world scenarios. More recent research\cite{ComiRec} utilizing multi-head attention to explore the potential of multi-interest routing that provide a richer representation of user interests. The exploration of users' multi-interests also remains a topic of active research in this field.\cite{li-etal-2022-miner, 10.1145/3604915.3608766, 9835441, 10.1145/3543873.3587341}

\section{Theory}

While there is no rigorous explanation as to why Tweedie is popular in actuarial science, it is generally considered that it closely captures the nature of claim events. Analogously, we observe a similar distribution pattern in user viewing time within recommendation systems, where negative samples far exceed positive ones due to users often making a single choice from the options presented. To further explain the intuition of applying this in predicting user-content viewing time, we can draw comparisons of hypothesis between recommendation system and actuarial science, offering more insights in table \ref{comparison}.

\begin{table}[ht]
	\caption{Comparison of mechanism between recommendation and actuarial science}
	\centering
	\begin{tabular}{ |p{1.7cm}||p{2.5cm}|p{3.0cm}|  }
		\hline
		Sub-process & Actuarial Science & recommendation \\
		\hline
		Poisson & claim frequency  & how many interests of each devices \\
		\hline	
		Gamma & claim severity  & the degree or intensity of each interests \\
		\hline
		Compound & what is the total cost of policy holder & what is the accumulated watching duration generated by each interests \\
		\hline
	\end{tabular}
	\label{comparison}
\end{table}

As is shown, the underlying intuition of Tweedie process is trying to consider the viewing time as the mixture behavior of one or more people with multiple interests. The hypothesis has more solid foundations in lots of recommendation system in terms of multi-interests, as is introduced in Section II. In reality it's a common case of shared devices used by multiple people especially for those over-the-top(OTT) devices. The presence of a range of genres within a single video \cite{SaurabhGenreSpec} caters to the diverse interests of multiple viewers during watching time. A more relevant work is done by Taejoon\cite{math12213365} who tried to use generalized linear model to predict user and item relevance using collaborative filter framework. 

\subsection{Analytical Comparison}

We try to compare the one variant of Tweedie losses in equation \ref{eq:2} with viewing-time-weighted Logloss\cite{DNYoutubeRec} mathematically,

\begin{equation}\label{eq:3}
	\begin{aligned}
		TLoss[\hat{y_i}] &= - \sum [y_i \frac{{\hat{y_i}}^{1-p}}{1-p} + \frac {{\hat{y_i}}^{2-p}} {2-p}] \\
		CLoss[\hat{y_i}] &= - \sum y_i [{y'}_i \ln{\hat{y_i}} + (1-{y'}_i) \ln{(1-\hat{y_i})}]
	\end{aligned}
\end{equation}
where $y_i$ is the target, raw viewing time, in Tweedie Loss. Meanwhile ${y}'_i \in \{0, 1\}$ is the label for classification in $CLoss$, the click event. 

The upcoming derivations will involve certain approximations.
Without loss of generality we can set $p = 1.5$ into equation \ref{eq:3}, in \ref{preanalysis} we will show that $p=1.5$ is our optimal choice for real world problem.
\begin{equation}
	\begin{aligned}
		TLoss &= - \sum {y_i} [- 2 {\hat{y_i}}^{- \frac{1}{2}} + 2 \frac{\hat{y_i}^{\frac{1}{2}}}{y_i} ] \\
	\end{aligned}
\end{equation}
Notice that $\hat{y_i}^{- \frac{1}{2}} = 1 + f(\hat{y_i}) + f(\hat{y_i})^2 + o(f(\hat{y_i})^3)$, where $f(\hat{y_i})=1-\sqrt(\hat{y_i})$ and further leverage Taylor series expansion of $\ln{x}$ for \textbf{positive samples} ($y_i = 1$), we can approximately derive individual terms in the formula of Tweedie loss and Logloss as:
\begin{equation}\label{eq:5}
	\begin{aligned}
		TLoss &\sim y_i [\frac{1}{1- (1 - \sqrt{\hat{y_i}})}] \\
		&\sim y_i [- f(\hat{y_i}) - f(\hat{y_i})^2 - o(f(\hat{y_i})^3)] \\
		CLoss &\sim {y'}_i \ln(\sqrt{\hat{y_i}}) = {y'}_i \ln[1-(1-\sqrt{\hat{y_i}})] \\
		&\sim {y'}_i [-f(\hat{y_i}) - \frac{f(\hat{y_i})^2}{2} - o(f(\hat{y_i}))^3)] \\
	\end{aligned}
\end{equation}

Obviously, when $\hat{y_i}<1$ and $\lim_{\hat{y_i} \to 1^{-}} \hat{y_i}$, Tweedie Loss is asymptotically more sensitive than Log Loss due to larger coefficients in higher order terms. Meanwhile $\hat{y_i}$'s domain is $[0, 1]$ under probability assumption of $CLoss$, suggesting that its predictive capacity on CTR is limited with compared to viewing time regression. For negative samples, similar reasoning can be applied. We argue that the Tweedie loss enables the model, represented as $\hat{y_i}=f(x)$, efficiently leverage the abundant information contained within features, $x$, in predicting viewing time. We believe qualitatively by this reason, Tweedie loss can works better than weighted LogLoss from monetization perspective. 

\subsection{Decomposition Framework}

In this session, we will build upon the theoretical framework outlined above to develop a method capable of decomposing various losses (including Tweedie loss and cross-entropy loss) and combining them into a unified loss function that accurately reflects our core business objectives.

If we have gathered some online results in terms of several compelling objectives, total viewing time, conversion, ctr, etc. We can consider decomposing different metrics in Hilbert space by projecting loss function into the sub-space of each specific metric. We believe that different subspaces represent distinct metrics, which together constitute a "whole" space. Our hypothesis is the projection coefficients based on one type of Taylor Space \cite{ZWICKNAGL201365} into sub-space for different metrics is the same. Taking metrics conversion and total viewing time as examples, the projection coefficients from loss functions to different metric subspace can be expressed as:
\begin{equation}\label{eq:1}
	\begin{aligned}
		\mathbf{Watch\_Duration} &= \vec{t} &= (w_{a1}, w_{b1}, w_{c1}) \\
		\mathbf{Coversion} &= \vec{v} &= (w_{a2}, w_{b2}, w_{c2}) \\
	\end{aligned}	
\end{equation}
And decomposing Loss functions into the 3rd order(it can be higher order) of one typical Taylor Series as in equations \ref{eq:5}. We get coordinates in $L_{g}$ as
\begin{equation}
	\begin{aligned}
		L_{g} &= \sum c_{i,g} f_i(x) \sim \sum (c_{1,g}, c_{2,g}, c_{3,g}) \cdot (f_1(x), f_2(x), f_3(x))^\intercal \\
	\end{aligned}
\end{equation}
where $f_n(x)=x^n$ is the basis power function, $g \in \mathbb G$ represent different loss functions accordingly, and assume $\overrightarrow{c_g}^\intercal =  (c_{1,g}, c_{2,g}, c_{3,g})$. Note that cardinality of $\mathbb G$ is $N$ for $N$ online experimental observations.

How to get the coefficient of $\vec{t}$ and $\vec{v}$? We believe there is always a gap between online business and our model loss, and we need a map from loss to our metric. Therefore, we conduct $N$ experiments independently to get those watch duration and conversion metric observations. Mathematically, we ended in $2N$ equations from $N$ online observation groups with 2 metrics each by projecting $L_{g}$ into different metric space, conversion and viewing time:
\begin{equation} \label{eq:4}
	\begin{aligned}
		\mathbf{Metric}_{\text{Watch\_Duration}, g} &= \overrightarrow{c_g}^\intercal \cdot \vec{t} \\
		\mathbf{Metric}_{\text{conversion}, g} &= \overrightarrow{c_g}^\intercal \cdot \vec{v} \\
	\end{aligned}
\end{equation}
Upon deriving solutions from the aforementioned equations, which could be an ordinary linear regression problem if $N$ is very large, we can get the projection coefficient $\vec{t}$ and can use that to formulate a new loss function. A practical strategy involves reconstituting a compounded loss parallel to the coefficient vector $\vec{t}$(for example), through a composite of the $N$ losses represented by $\overrightarrow{c_g}$. This compound loss function is designed specifically to maximize a single objective, like total viewing time. 

\section{Experiments}

In this section, we present two types of experiments to evaluate the effectiveness of our proposed approach. First, we perform a user simulation study on synthetic data, which serves two key purposes: (i) ensuring reproducibility, and (ii) providing a basis for future research to build upon. To facilitate further investigations, we make our user simulation source code publicly available.\footnote{https://github.com/fucusy/Tweedie-Regression-for-Video-Recommendation-System} Second, we demonstrate the real-world applicability of our method by conducting experiments on industrial data, including both offline pre-analyses and online A/B tests. These complementary results offer compelling evidence that the proposed method is effective under real-world conditions. Specifically,  we model normalized viewing time by Tweedie regression.

\subsection{User Simulation on Synthetic Data}

In order to conduct a controlled study that closely mirrors real-world behavior while remaining fully reproducible, we implement a synthetic user simulation. Our focus is on new-user recommendations, which allows us to simplify certain assumptions while still capturing key dynamics of user--item interactions. 

\subsubsection{Simulation Assumptions} 

\begin{itemize}
    \item Each title has a static click probability, drawn from a normal distribution.
    \item Users are featureless (no user-level features are modeled).
    \item Once a user clicks on a title, they watch a portion of the content before deciding whether they want to watch until the end .
    \item Each title has an inherent probability of being completed (``completion intention''), also drawn from a normal distribution.
    \item Titles have two possible completion rate distributions, corresponding to users who intend to finish and those who do not.
    \item Each title has a unique duration sampled from a normal distribution.
    \item Users scroll through a ranked list of titles from the beginning, with a certain probability of stopping at any point. This behavior simulates an abandonment event where the user opts not to view subsequent titles in the list.
\end{itemize}

\subsubsection{Simulation Protocol and Settings}

We simulate a 13-day scenario. During the first 3 days, users are served a human-edited ranking, and their interaction data are collected and accessible for models. Beginning on day 4 and continuing through day 13, our model generates the ranking, while newly collected user feedback data are iteratively incorporated into the training process. By updating the model daily with fresh data, we capture temporal dynamics and assess how the model adapts over time.

We simulate 10{,}000 users and 1{,}000 titles. For each user, we record whether they click on a title, how long they watch. 

All models are trained on the collected data samples (i.e., interactions with titles that users actually see) using stochastic gradient descent with a learning rate of \(1\times10^{-3}\). We train each model for 100 epochs, shuffling the data to mitigate potential biases from ordering effects.

\subsubsection{Model Architectures and Loss Functions}
We compare four models that share a common neural architecture but differ in their loss functions.

Each title ID is first mapped to a 16-dimensional embedding. This embedding is then passed through one fully connected layer (mapping from 16 to 8 dimensions) followed by another layer that outputs a scalar. Formally:
\[
\text{title\_embedding} \in \mathbb{R}^{16} 
\,\xrightarrow{\text{FC}_1}\,
\mathbb{R}^{8}
\,\xrightarrow{\text{FC}_2}\,
\mathbb{R}.
\]
This scalar is interpreted differently for each loss function.

\noindent\textbf{Pointwise Model (Click Optimization).} 
We optimize the click probability using a standard logistic loss:

\[
LogLoss[\hat{y_i}] = - [{y'}_i \ln{\hat{y_i}} + (1-{y'}_i) \ln{(1-\hat{y_i})}]
\]

\noindent\textbf{Watch-Duration Weighted Model.}
We again optimize for click probability but weight each clicked sample by its total watch duration:
\[
WeightLogLoss[\hat{y_i}] = -y_i [{y'}_i \ln{\hat{y_i}} + (1-{y'}_i) \ln{(1-\hat{y_i})}]
\]

\noindent\textbf{Regression Model.}
Here, we replace the classification-based loss with mean-squared error (MSE) to predict a continuous outcome (e.g., total watch time):
\[
MeanSquaredError[\hat{y_i}] = (\hat{y_i} - y_i)^2.
\]

\noindent\textbf{Tweedie Regression Model.}
We employ a Tweedie distribution-based loss function to accommodate the skewed nature of watch durations. $p$ is set to 1.5 for verification. 
\[
TweedieLoss[\hat{y_i}] = - y_i \frac{{\hat{y_i}}^{1-p}}{1-p} + \frac {{\hat{y_i}}^{2-p}} {2-p} \\
\]

\subsubsection{Simulation Results}

\begin{figure}[h]
	\label{watch_duration_reward}
	\centering
	\includegraphics[width=0.5\textwidth]{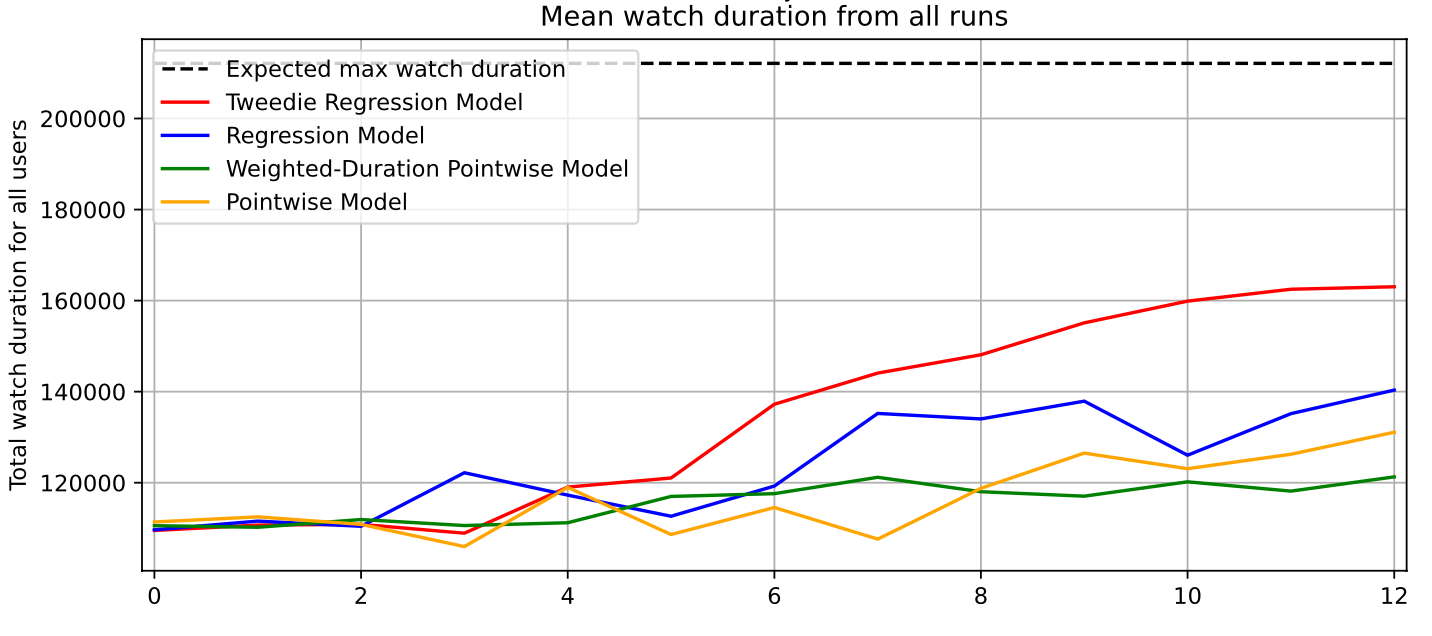}
	\small
	\caption{
	A simulation of users' watch time was conducted using different loss function models, each model run multiple times. This represents the average watch duration across all runs for each model over time.}
\end{figure}

To ensure the stability of model performance, each of the four models was run 10 times, and the mean watch duration from each run is shown in Figure~\ref{watch_duration_reward}. Among all tested models, the Tweedie Model achieves a higher mean watch duration compared to the Regression Model, the Weighted-Duration Pointwise Model, and the Pointwise Model.

For a more comprehensive comparison, for the collected 10 total watch duration reward data points for each model (excluding the first 3 days, which relied on manually edited rankings). These data points form the basis of our statistical analysis, shown in Table~\ref{significance}. The Tweedie Model exhibits higher total watch duration rewards relative to the other models, with statistically significant performance lifts as evidenced by the p-values.

\begin{table}[ht]
	\caption{Total watch duration reward comparision: significance calculation}
	\centering
	\begin{tabular}{ |p{3.5 cm}||p{1.2cm}|p{1.0cm}|  p{1.0cm}| }
		\hline
		Models           & \textbf{Regression} & \textbf{Weighted} & \textbf{Pointwise} \\
		\hline
		\textbf{Tweedie duration reward lift}   & +10\%& +21\%  & +20\% \\
		\hline		
		 \textbf{Lift p-value}   &  0.005 &  0.0006 & 1.7e-5 \\
		\hline
	\end{tabular}
	\label{significance}
\end{table}

\subsection{Pre-analysis of Real-world Application}\label{preanalysis}

Our aforementioned hypotheses on multi-interests are challenging to verify directly, as we do not obtain detailed data regarding the number of viewers behind each device screen. Furthermore, obtaining comprehensive insights into each individual's latent interests, with respect to intensity or degree, is impractical. Luckily we can observe the similarity between distribution of viewing time from user behaviors on our platform and the Tweedie distribution. First, we compare the ratio of number of positive and negative samples based on raw viewing time. Here, negative samples indicate instances where users did not engage with the content that was presented to them(watch time = 0). And by fine-tuning hyper-parameter $p$ of Tweedie distribution, we can easily achieve approximately equal ratio from real world distribution to ideal Tweedie distribution. 

Second, we apply Z-score normalization method on raw viewing time and then tried to truncate the viewing time at some fixed upper bound. The ultimate distribution of viewing times further exhibits patterns analogous to the Tweedie distribution, characterized by a pronounced peak at zero and a more gradual peak at a value greater than zero. As mentioned before, Figure\ref{normalized_vt} shows the distribution of our Z-score normalized real viewing time data. We also do a grid search for finding the optimum of Komogorov-Smirnov-statistics by varying our Tweedie distribution parameters, $\mu$, $p$ and $\phi$. The optimal hyper-parameter of Tweedie distribution when achieved minimized KS statistic at around 0.05, are $t \sim 0.05$, $\mu \sim 0.2$, $p \sim 1.5$, $\phi \sim 1.5$. Although it's not sufficient to accept the null Komogorov-Smirnov test hypothesis, the optimal parameter $p$ demonstrates that the Tweedie distribution, which follows a compound Poisson-gamma process, offers the best fit for replicating real-world viewing behavior within the family of exponential dispersion models. Since $p=1.5$ gives us the lowest KS statistics, in our theoretical analysis and experiment setup, we choose 1.5 as our p value in Tweedie loss.

\subsection{Real-world Application Online Experiment}

Our company carried out A/B testing for the aforementioned treatment using our internal experimental system. 

The training dataset of ranks contains several hundred million samples, which are derived from several million unique viewers who engaged in a week's online presentation events in our internal system. Both the control and treatment groups utilize an identical training set in terms of raw features, size, and other characteristics. Each user-item pair has a few hundred features both numerical and categorical. There are plenty of feature engineering processes, as well as data cleaning for getting the true valuable labels. All those processes are the same for control and treatment. The control group we are talking about is the best strategy we get so far for our recommendation system in terms of optimizing viewing time. Before that, we have a series of experiments, including list-wise, pair-wise ranking, different data cleaning strategies, different weights strategies, etc. Till the beginning of this experiment, we are reaching the best solution which is point-wise ranking with viewing time as weights.

Other differences except the loss function: the control model treated video viewing time as sample weights during training, while the treatment model used an equal weighting approach in the training dataset. Table \ref{online} shows the relative lift during online experiments for 1 week.

\begin{table}[ht]
	\caption{Online evaluation metrics}
	\centering
	\begin{tabular}{ |p{1.6cm}||p{1.5cm}|p{1.5cm}|p{2.2cm}|  }
		\hline
		Exp Group & conversion 5-min & total viewing time & Revenue \\
		\hline
		Control(C)   & --    & -- & -- \\
		Treatment & -- 0.17\% & + 0.15\% & + 0.4\% \\
		\hline
	\end{tabular}
	\label{online}
\end{table}

The online results of Tweedie treatment show 0.4\% significant lift in revenue, 0.15\% significant lift in device average viewing time, when applying Tweedie regression on video watch time. Conversely, there was also a significant decline of 0.17\% in the conversion rate (defined as the percentage of users who watched content on our platform for more than 5 minutes during the experiment period). Such divergent behavior in key metrics is uncommon in typical experiments, where viewing time and conversion rate usually rise or fall together. Still, these findings are in accord with our hypothesis that optimizing for revenue or viewing time might require a trade-off in conversion, which correlates more closely with CTR. 

\section{Conclusion \& Future Work}

Our work endeavors to bridge the gap between true business goals and the ranking objectives within video recommendation systems by transitioning the ranker's loss function to the Tweedie loss. To the best of our knowledge, this is the initial attempt regard ranking as Tweedie regression problem. 
Our analysis discuss the underlying mechanics of user behavior by Tweedie process, providing evidence that the Tweedie loss emerges as the prime candidate within the exponential dispersion model family for our regression tasks. Further, we illustrate the theoretical proximity of the Tweedie loss function to Logloss and discuss why Tweedie model yields superior results. Additionally, we propose a strategy that aim at optimization towards a singular objective.

To demonstrate its effectiveness, our simulation results on synthetic data clearly demonstrate that Tweedie loss significantly outperformed both ordinary regression and classification loss with or without weights. Furthermore, the outcomes from our online experiments verified that Tweedie loss is aligned with our true business goal—enhancing revenue and total viewing time—while resulting an acceptable decrease in conversion rates. The experiment also suggests that Tweedie loss are moving the system towards maximizing watch time instead of CTR. 

Our ongoing research will probe various methods to reconcile the objectives of recommendation models and the actual business goals. Efforts will focus on novel approach of modeling that better match business targets, applying Mixture-of-Expert architecture on multi-task objectives under decomposition framework, or integrating ranking loss into the optimization process located in corresponding online advertising.

\bibliographystyle{IEEEtran}
\bibliography{myref}
\end{document}